\documentclass{article}

\usepackage[final]{iai_neurips_2024}

\usepackage[utf8]{inputenc} %
\usepackage[T1]{fontenc}    %
\usepackage{hyperref}       %
\usepackage{url}            %
\usepackage{booktabs}       %
\usepackage{amsfonts}       %
\usepackage{nicefrac}       %
\usepackage{microtype}      %

\usepackage{graphicx}
\usepackage{subfigure}
\usepackage[capitalize,noabbrev]{cleveref}
\usepackage{wrapfig}

\usepackage{multirow}
\usepackage{multicol}

\usepackage[dvipsnames]{xcolor}
\definecolor{mydarkblue}{rgb}{0,0.08,0.45}

\hypersetup{
    colorlinks=true,
    linkcolor=mydarkblue,
    citecolor=mydarkblue,
    filecolor=mydarkblue,
    urlcolor=mydarkblue,
}

\newcommand{\noncritical}{auxiliary}

\usepackage{cvpr_macros}

\title{How Do Training Methods Influence the\\Utilization of Vision Models?}

\author{%
    Paul Gavrikov$^{1,2}$ \qquad Shashank Agnihotri$^{2}$ \qquad Margret Keuper$^{2,3}$ \qquad Janis Keuper$^{1,2}$\\\\
  $^{1}$ IMLA, Offenburg University, Germany \\
  $^{2}$ University of Mannheim, Germany \\
  $^{3}$ Max-Planck-Institute for Informatics, Saarland Informatics Campus, Germany \\
}

\begin{document}

\maketitle

\begin{abstract}
Not all learnable parameters (\eg, weights) contribute equally to a neural network’s decision function. In fact, entire layers’ parameters can sometimes be reset to random values with little to no impact on the model’s decisions.
We revisit earlier studies that examined how architecture and task complexity influence this phenomenon and ask: \textit{is this phenomenon also affected by how we train the model?}

We conducted experimental evaluations on a diverse set of ImageNet-1k classification models to explore this, keeping the architecture and training data constant but varying the training pipeline. 
Our findings reveal that the training method strongly influences which layers become critical to the decision function for a given task. 
For example, improved training regimes and self-supervised training increase the importance of early layers while significantly under-utilizing deeper layers. 
In contrast, methods such as adversarial training display an opposite trend.
Our preliminary results extend previous findings, offering a more nuanced understanding of the inner mechanics of neural networks.

\textbf{Code:} \url{https://github.com/paulgavrikov/layer_criticality}
\end{abstract}

\begin{figure*}[]
    \centering
    \rotatebox{270}{\includegraphics[height=0.85\linewidth]{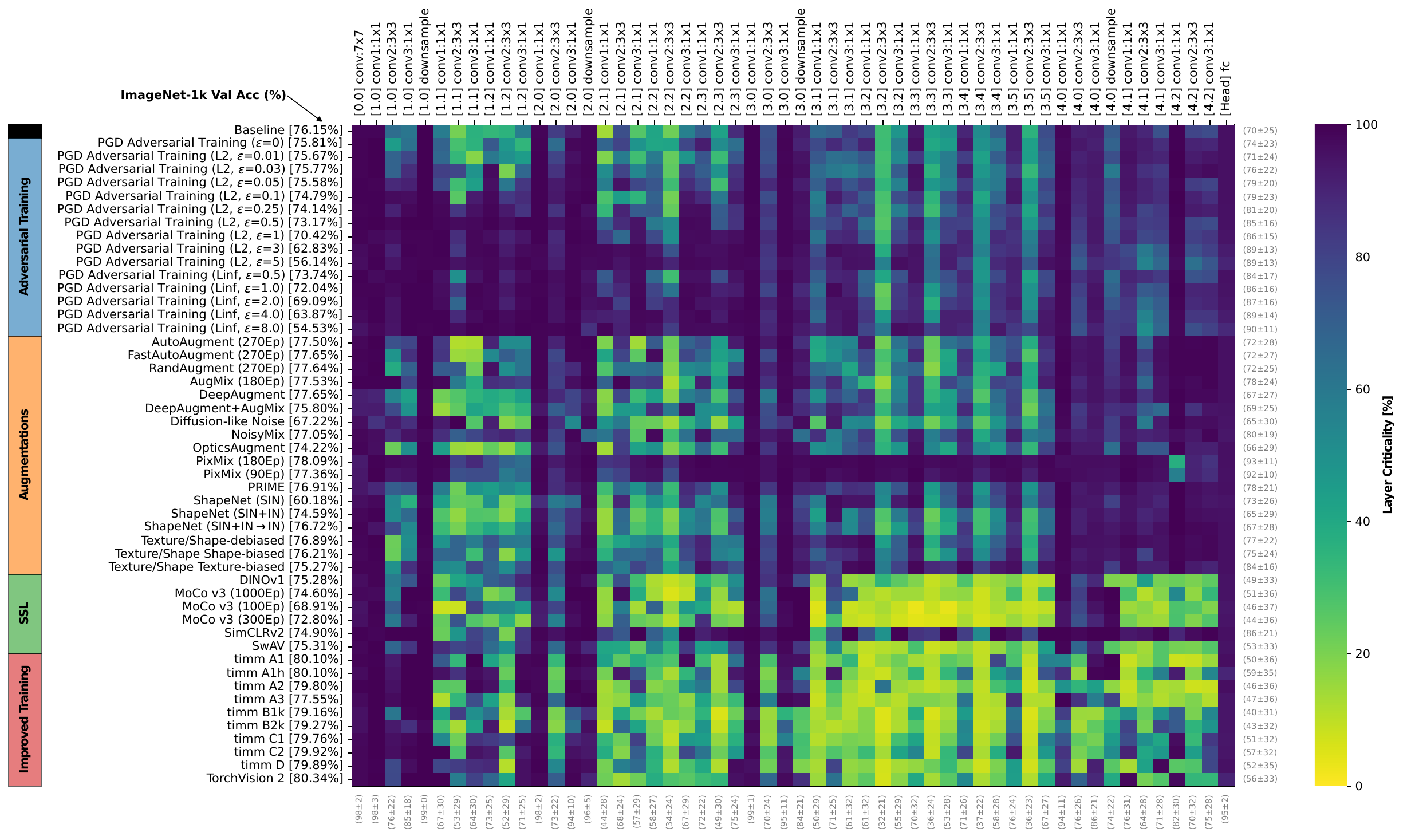}}
    \caption{\textbf{Training methods determine what layers become critical.} We measure the criticality of fifty different ResNet-50-based models that all utilize the same exact network architecture and training data (ImageNet-1k) but differ in their training methods. Darker spots denote layers that are \textit{critical}, \ie, in significantly different predictions and decreased performance after reset. Brighter spots are \textit{\noncritical}, \ie, resetting these layers does not significantly affect the model. We denote the average (mean$\pm$std) layer criticality for both, a model across layers on the right, for a layer across model on the bottom.}
    \label{fig:weight_reset_r50_training}
\end{figure*}
\section{Introduction}

A famous neurology myth often misattributed to Albert Einstein states that humans only use $10\%$ of the neural connections in their brains \citep{Radford1999Apr}. 
While modern research assumes that humans use all neural connections \citep{Boyd2008Feb,Herculano-Houzel2009} -- the same cannot be said about artificial neural networks. 
Quite the contrary, it is well known that trained neural networks do not utilize their entire capacity. 
This becomes evident through the lens of \textit{parameter pruning} \citep{lecun1989braindamage,hassibi1993brainsurgeon}, where (large numbers) of neurons can be removed after training without affecting performance, or, alternatively, the \textit{distillation} of large into equivalent smaller networks \citep{hinton2015kd,hoffmann2021towards}. 
Alas, the learned decision function only occupies a fraction of the neural network and the remaining neurons seem to be wasted. 

\citet{zhang2022arealllayers} showed that this seems to affect layers disproportionally. 
The learnable parameters\footnote{Throughout the rest of the article, the term \textit{parameters} specifically refers to learnable parameters.} of some layers are \textit{critical} to the decision function and replacing them with any other values than the learned ones (significantly) affects accuracy. 
In contrast, the performance is barely affected when the parameters in \textit{\noncritical}~layers are randomized. For instance, entire residual blocks of \textit{ResNets} \citep{resnet} trained on ImageNet \citep{imagenet} can be randomized without hurting accuracy. 
Affected layers seem to be dictated by training data size or more generally the complexity of the training function in addition to the architecture. 
\citet{Chatterji2020The} have even extended these findings to generalization, by showing that the average module criticality correlates with performance - \ie, more complex networks seem to generalize better.

We revisit the prior findings of \citep{zhang2022arealllayers,Chatterji2020The} on image classification models, which were obtained under rather clean conditions, such as the absence of weight decay \citep{Krogh1991} or batch normalization layers \citep{ioffe15} during training, and an overall simple training pipeline. 
These conditions do not reflect practical training pipelines well.
Thus, we raise the question: \textit{how does the training method affect layer criticality?}
To this end, we study criticality on a large model zoo of image classification models where the training data and architecture are fixed but the training pipeline is modified. 

Specifically, we use the model zoo of \citep{gavrikov2024biases} with a few replacements and select \textit{fifty} different ResNet-50 image classification models \citep{resnet} that all utilize the exact same network architecture but were trained in different manners. Our changes to training belong to the following categories: a \textit{baseline} model following the original training \citep{resnet};  various \textit{augmentation} techniques \citep{hendrycks2020augmix,hendrycks2022pixmix,hendrycks2021rendition,jaini2023intriguing,Muller_2023_ICCV,modos2022prime,li2021shapetexture,erichson2022noisymix,Cubuk_2019_CVPRAutoAugment,Cubuk2020RandAugment,Lim2019FastAutoAugment,geirhos2018imagenettrained}; \textit{adversarial training} against a projected gradient descent (PGD) adversary \citep{madry2018towards,salman2020transfer} -- which technically, is also a form of augmentation, but we observed significantly different behavior of these models; various \textit{self-supervised learning (SSL)} approaches \citep{chen2020simclrv2,Chen_2021_ICCV,caron2021emerging,caron2020swav} with supervised finetuning of the classification-head; and improved \textit{training recipes} in \textit{timm} \citep{rw2019timm,wightman2021resnet} and \textit{PyTorch} \citep{PyTorch,BibEntry2023Nov} combining multiple hyper-parameter optimizations into supervised training.

\section{Methodology}

Our study aims to assess the contribution of individual layers to a neural network’s decision function. To gauge a layer’s importance, we replace its parameters with random values (\textit{randomization}). If the network’s decisions remain largely unchanged after randomization, it suggests that the learned parameters contribute little beyond noise. 

This methodology largely follows \citet{zhang2022arealllayers}, who reset the parameters of individual layers to values drawn from the original initialization\footnote{We match the initialization used by each training method.}.
However, while \citet{zhang2022arealllayers} measured \textit{criticality} of a layer by the change in accuracy due to the randomization, we measure the angle between the probability vectors resulting from the randomization. 
Specifically, we apply a Softmax function to the network logits to obtain (pseudo-)probabilities, measure the cosine distance between those before and after randomization, and aggregate the measurements into a single scalar by averaging over all samples. 
The effect of each layer randomization on this measured distance is what we define as the \textit{criticality} of a layer.

This methodological change can be evaluated in an {unsupervised} manner and more importantly, is also sensitive to changes in the probability distribution including variations in errors (we refer the reader to \citep{Geirhos2020error} for a discussion on why this is important). 
As such, it provides a more holistic measurement of consistency in the decision before and after randomization well beyond correct predictions.

We call a layer \textit{\noncritical} if the decision is insignificantly affected by the reset ($\approx 0\%$ criticality) and \textit{critical} ($\approx 100\%$ criticality) if the distance between decisions changes significantly. 
Realistically, the criticality for most layers does not lie on the extremes of this spectrum, but anywhere in between. %
Due to a significant variance (standard error of up to $45\%$ on a few layers in specific models; see \cref{fig:weight_reset_r50_training_std}) 
in criticality on some layers, we repeat experiments with different random seeds and report the mean over three trials.
For computational reasons, we evaluate layers on a subset of 10,000 random images from the ImageNet ILSVRC-2012 challenge validation set \citep{imagenet}.

\citet{zhang2022arealllayers} analyzed residual blocks as a whole -- in contrast, we more meticulously randomize individual layers, which include different convolution and fully-connected layers. 
However, we do not re-initialize batch normalization layers \citep{ioffe15} to avoid signal propagation issues.

\section{Results} 
Due to the wide use and availability of pre-trained models, currently, all our results are obtained on ResNet-50 \citep{resnet}. 
Recall that this architecture consists of a stem (denoted by \textit{[0.*]}), 4 stages (denoted by \textit{[1-4.*]}), a pooling layer, and a fully-connected classification head (denoted by [Head]). 
Each stage consists of several \textit{residual} bottle-neck blocks which include learnable $1\times 1$ convolutions (\textit{conv1}, \textit{conv3}), $3\times 3$ convolutions (\textit{conv2}), as well as batch-normalization layers \citep{ioffe15}. The first residual block in each stage is special, as it downsamples by a strided convolution, thus, adding a learnable layer on the skip connection (\textit{downsample}).

\textbf{General Observations.} The results in \cref{fig:weight_reset_r50_training} (analogously see \cref{sec:appendix:additional} for more views) clearly show that the training method influences what layers become critical -- despite that all models were trained on the same training set (some with more extreme forms of data augmentation utilizing a negligible amount of extra data). 

In contrast to previous findings \citep{zhang2022arealllayers,Chatterji2020The}, we observe that \textbf{no layer is always \noncritical} across training methods. For instance, we observe an average criticality of just $36\%$ for a spatial convolution layer (\textit{[3.5] conv2}). 
Yet, if we randomize the same layer in a \textit{PixMix} \citep{hendrycks2022pixmix} model, we observe a strong criticality of $95\%$. 
On the opposite, we do find layers that are always critical. As expected, these include the initial stem convolution (\textit{[0.0] conv}) and the classification head (\textit{fc}). 
Beyond, we find that most first convolution layers in each stage (\textit{[*.0] conv1}) are critical -- yet the number of outliers increases with depth. 
Similarly, we find that the downsampling convolution (\textit{[*.0] downsample}) in each stage is often critical. 
In stage 1 this layer is critical for all models but again the criticality of deeper downsampling convolutions depends on the training strategy used for the model. 
Lastly, akin to \citet{gavrikov2023power}, we find that pointwise convolution layers tend to be more critical than spatial convolution layers (except for the stem). For all other layers, criticality depends on the training method. In the following paragraphs, we analyze specific categories of training methods.

\textbf{Adversarial Training (AT).} This training technique intends to increase the robustness of neural networks by training on adversarially perturbed training samples \citep{madry2018towards}. 
To avoid perturbations that cause a shift in semantic meaning, perturbations are often constrained by an attack perturbation budget $\epsilon$ for some $\ell_p$ norm.
We consider AT using a PGD attack \citep{madry2018towards} which optimizes the perturbations over several iterations (here: three). Please note, that reported $\epsilon$ values for $\ell_\infty$ norms are short for $\epsilon / 255$ (but not for the $\ell_2$ norm).

\begin{figure}[h!]
    \centering
    \includegraphics[width=0.7\linewidth]{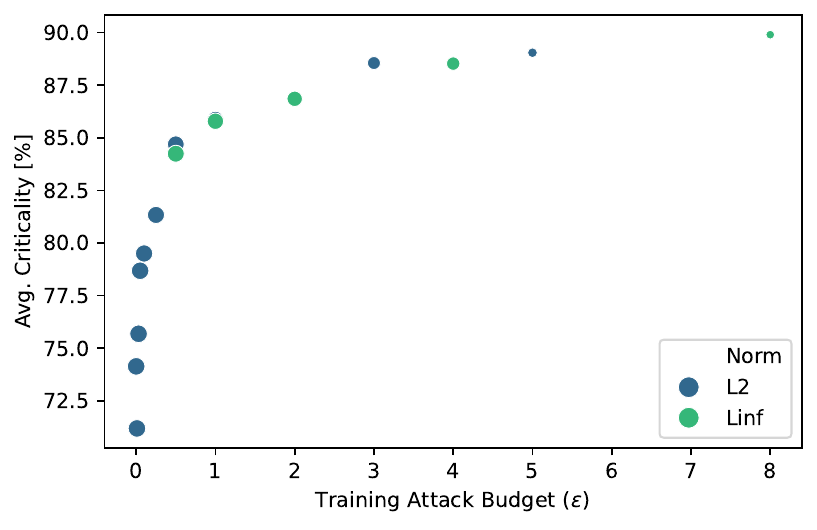}
    \caption{\textbf{Adversarial training increases the average criticality proportional to the training attack budget $\epsilon$.} We ablate $\ell_\infty$ from $\ell_2$-norm training but do not observe any significant differences in their trends. The marker size in the plot indicates the validation accuracy on ImageNet-1k (larger is better).}
    \label{fig:weight_reset_adv_training}
\end{figure}

We find that AT increases the criticality proportional to the attack budget $\epsilon$ during training results. To make this more tangible, we average the criticality over all layers and show the results in \cref{fig:weight_reset_adv_training}. 
We do not observe differences between training that utilizes $\ell_2$ or $\ell_\infty$ norms for attacks. 
Our findings in \cref{fig:weight_reset_r50_training} suggest that neural networks utilize more of their capacity under increasing training attack strength. 
This augments previous findings that showed similar insights through accuracy improvements with larger networks \citep{madry2018towards} or richer representations in convolutions filters \citep{Gavrikov_2022_CVPR,Gavrikov_2022_CVPRW}. 
AT primarily increases the criticality in the layers of the first and second stages and slightly in the third stage. The criticality of layers in the fourth stage is barely affected but rather decreases, compared to the baseline (please refer to \cref{fig:weight_reset_r50_training_delta}).

\textbf{Augmentations.} 
Compared to AT, the influence of different augmentation strategies seems weaker. We do find that augmentations tend to increase the average criticality, \ie, they do occupy more of the network capacity -- but most changes are rather small. Affected layers seem to fluctuate by method, but we find that all augmentation methods consistently increase the criticality of some of the deepest layers (\textit{[4.0] downsample, [4.1] conv2, [4.2] conv2/3}). Similar to the organization of the human brain \citep{hubelwiesel1979brainmech}, deeper layers in neural networks are associated with activations of more complex features. For images, these tend to correspond to shapes as opposed to texture information that is captured by early layers \citep{lecun1989cnn,yosinski2014howtransferable}. Indeed, prior work has observed that our tested augmentations increase in their shape responses \citep{gavrikov2024biases}. Thus, a reasonable hypothesis is that the increased criticality in deeper layers correlates with stronger shape representations.

Strong outliers to our observations are the \textit{PixMix} models \citep{hendrycks2022pixmix}. These models have the highest average criticality in our model zoo without a single \noncritical~layer. The augmentation technique has been shown to improve multiple safety dimensions beyond test accuracy and combined with the findings of \citet{Chatterji2020The} it may indeed suggest that a higher degree of criticality correlates with ``better'' neural networks.

\textbf{Self-Supervised Learning (SSL).}
SSL has been shown to produce rich representations for many downstream tasks (see \citet{oquab2024dinov} for a recent example) as the granularity and implicit biases of annotations do not confine it. 
Interestingly, we find that \textit{DINO} \citep{caron2021emerging}, \textit{MoCo v3} \citep{Chen_2021_ICCV}, and \textit{SwAV} \citep{caron2020swav} severely differ in their criticality measurements from the supervised models we discussed before. These SSL models have a large presence of \noncritical~layers in the last three stages but (slightly) increase in criticality of early layers. This suggests that SSL learns shorter decision functions and, thus, seeks to strengthen early operations. However, we again find an outlier: \textit{SimCLRv2} \citep{chen2020simclrv2} has no \noncritical~layers and shows a somewhat similar distribution of criticality to \textit{PixMix}.

\textbf{Improved Training Recipes.} The standard 90 epoch ImageNet training with simple augmentations was shown to be suboptimal for many models including ResNets \citep{wightman2021resnet}. Modern training recipes utilize a significantly more complex set of training tweaks -- often in combination with longer training schedules \citep{wightman2021resnet,BibEntry2023Nov}. Similar to our findings on self-supervised learners, these improved training methods appear to shift model decisions to early operations while relaxing deeper operations. Most notably, this increases the criticality of the first and second residual blocks. These models show great improvements in generalization on many datasets but at the same time, they even further prioritize texture information \citep{gavrikov2024biases}. Ultimately, this may suggest that these techniques better fit ImageNet problems but may not provide improvements for representations beyond.

\begin{figure}[!]
    \centering
    \includegraphics[width=0.7\linewidth]{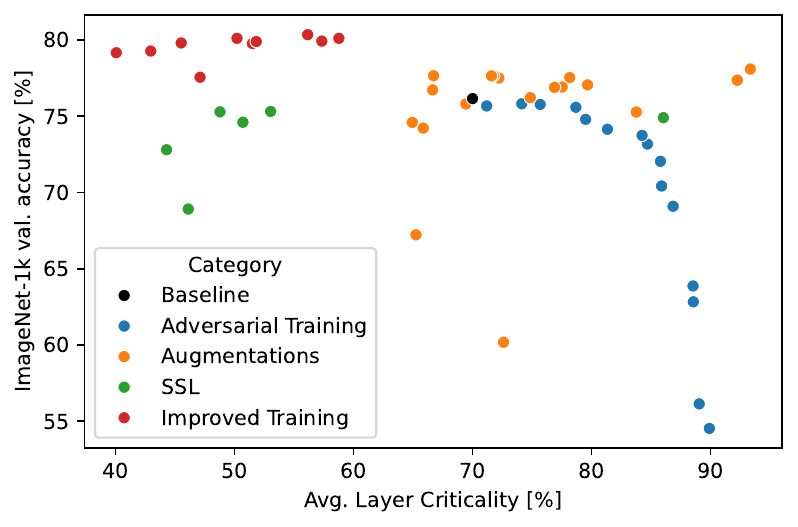}
    \caption{\textbf{Correlation between average network criticality and performance on ImageNet-1k.}}
    \label{fig:acc_vs_crit}
\end{figure}

Lastly, we attempt to correlate average layer criticality with accuracy on ImageNet-1k, across different training strategies in \cref{fig:acc_vs_crit}.
Ignoring the category labels we observe a moderate Spearman's $R=-0.46$, but when we remove the adversarially trained models the correlation is faint (Spearman's $R=-0.17$). 
Thus, there is a low likelihood of a causal connection between ImageNet accuracy and layer criticality.
\section{Conclusion, Limitations, and Future Work}

Our ongoing study extends previous findings about the complexity of learned decision functions of image classification models.
Instead of analyzing individual models as often done in mechanistic interpretability works (\eg, \citep{Olah2020,goh2021multimodal}) we focus on the common impact of training methods on layer criticality. 
We have shown that some forms of training leave distinct patterns in the decision function. 

\paragraph{Discussion.} 
Assessing the generalization of neural networks through benchmarks can be tricky, as models may specialize in specific settings at a cost to others. Capturing all of these nuances in static test datasets is often an unrealistic journey. For example, models excel in accuracy on the ``clean'' ImageNet data but drop in performance if the same samples are corrupted \citep{hendrycks2019commoncorruptions}. In that sense, the decision function complexity may offer a better (relative) assessment of generalization by focusing on the inner mechanics of models as opposed to benchmark results. Yet, it remains to be seen if the observed correlations of \citet{Chatterji2020The} hold on larger model zoos and wider notions of generalization beyond clean accuracy on the ImageNet validation set.

However, even if criticality and generalization are orthogonal to each other, the complexity of the decision function may still be relevant and offer a better explanation of phenomena that were linked to (adversarial) robustness before. For instance, these include more human-likeness \citep{NEURIPS2021_c8877cff,Gavrikov_2023_CVPRW}, better calibration \citep{grabinski2022overconf}, and transferability \citep{salman2020transfer} under adversarial training. \citep{Gavrikov_2022_CVPR,Gavrikov_2022_CVPRW} have shown that robust models contain more diverse feature representations and suggested that this may be linked to transferability.

\paragraph{Future Work.} 
We aim to extend our study to understand if the distinct patterns in the decision function patterns may be an artifact of the scenario we studied. 
For instance, we wonder if layer criticality depends on the test data - \ie, what happens if we replace ImageNet test data with corrupted images \citep{hendrycks2019commoncorruptions}, different renditions \citep{hendrycks2021rendition}, or adversarial perturbations \citep{goodfellow2015explaining}. 
Another dimension to explore is the architecture -- we are curious if our findings scale to modern network classification architectures such as \textit{ConvNeXt} \citep{liu2022convnet_convnext}, \textit{vision transformers (ViTs)} \citep{dosovitskiy2021an}, zero-shot classification with join-embedding models like \textit{CLIP} \citep{radford21clip}, (classification) prompting on large language models with vision capabilities such as \textit{BLIP} \citep{li2022blip}. 
How does criticality change under architectural interventions aimed at robustness improvements, \eg, \citep{grabinski2022frequencylowcut,lukasik2023improving,agnihotri2023unreasonable,agnihotri2024beware,agnihotri2024improvingfeaturestabilityupsampling} and their task-specific robustness tests \citep{cospgdagnihotri24b}?
Can we maybe even use this method to better understand the symbiotic connections of individual modalities in multi-modal models parallel to the approaches in \citep{goh2021multimodal,gavrikov2024vision}?

Beyond mechanistic interpretability, we also hope that our results could guide practical applications such as model compression and transfer learning. 
We would expect, that \noncritical~layers do not need to be distilled, or alternatively could be pruned without significantly affecting performance. 
As such, compression techniques may be affected by the training method. 
Additionally, it was shown that robust models perform better in transfer learning \citep{salman2020transfer} but is this indeed due to the increased robustness? 
We have observed that robust models also have a high ratio of critical layers, perhaps this might be a better explanation.

\paragraph{Acknowledgements.} S.A. and M.K. acknowledge support by the DFG Research Unit 5336 - Learning to Sense (L2S).

\bibliography{main}
\bibliographystyle{icml2024}

\newpage
\appendix

\section{Additional Analysis}
\label{sec:appendix:additional}
\begin{figure}[h]
    \centering
    \includegraphics[width=1.0\linewidth]{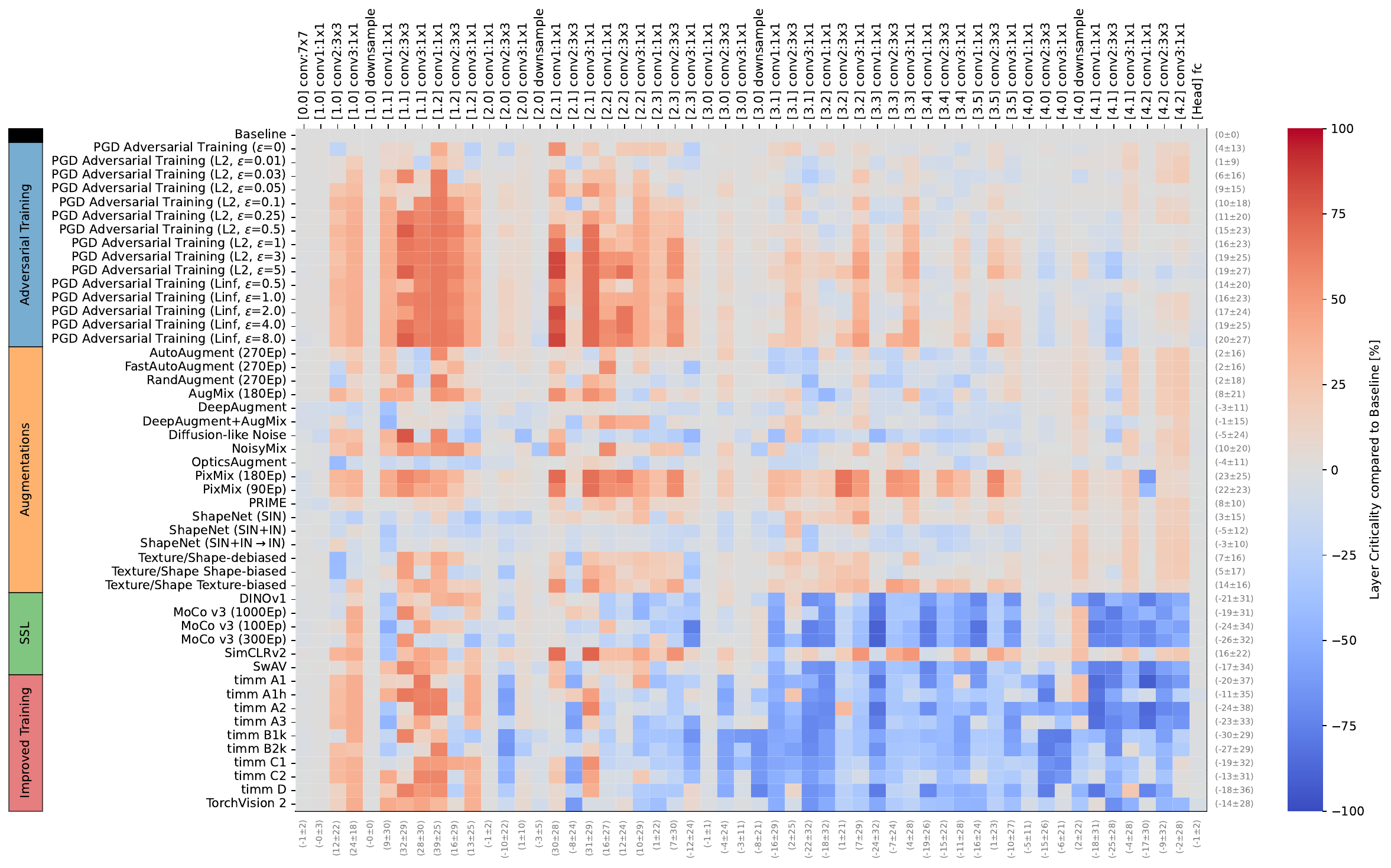}
    \caption{\textbf{Criticality difference to the baseline.} In addition to the plot in \cref{fig:weight_reset_r50_training}, we here show the difference to the baseline model \citep{resnet}. Positive numbers indicate increases in criticality and negative numbers decrease.}
    \label{fig:weight_reset_r50_training_delta}
\end{figure}

\begin{figure}[h]
    \centering
    \includegraphics[width=1.0\linewidth]{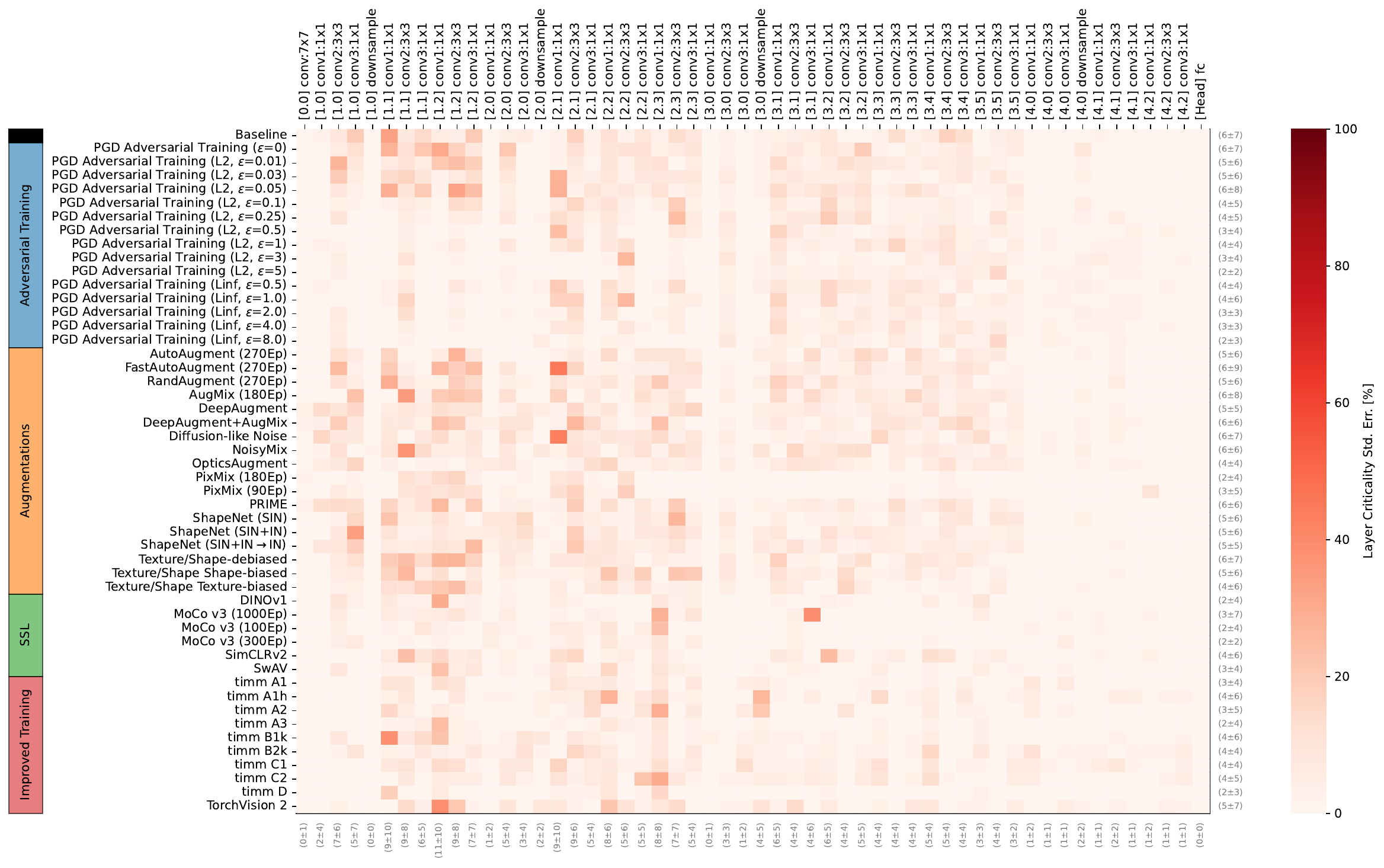}
    \caption{\textbf{Criticality standard error.} In addition to the plot in \cref{fig:weight_reset_r50_training}, we here show the standard error in criticality measurements over 3 runs with different seeds for the randomization.}
    \label{fig:weight_reset_r50_training_std}
\end{figure}

We alternatively visualize the results from \cref{fig:weight_reset_r50_training} by plotting the difference in the criticality of each layer with respect to the baseline in \cref{fig:weight_reset_r50_training_delta}. \Cref{fig:weight_reset_r50_training_std} displays the standard error of measurements over three independent runs with different seeds for the randomization.


\newpage
\section{Model Training Strategy Details}
In \cref{supp_tab:performance}, we provide a legend for the training strategies considered.
\begin{table}[!h]
\centering
\small
\caption{An overview of the utilized models (training strategies) in our study. }
\label{supp_tab:performance}
\scalebox{1.00}{
\begin{tabular}{llr} 
\toprule
&\multicolumn{1}{l}{\textbf{Model} }                                                                            & \multicolumn{1}{c}{\textbf{ImageNet accuracy} [\%]}                    \\ 
\midrule
             & Original Baseline \citep{resnet}                                                   & 76.15                 \\ 
\midrule
\multirow{18}{*}{\rotatebox{90}{\textbf{Adversarial Training}}}& PGD-AT ($\epsilon$=0) \citep{salman2020transfer,madry2018towards}        & 75.81                 \\
& PGD-AT ($\ell_2$, $\epsilon$=0.01) \citep{salman2020transfer,madry2018towards}     & 75.67                 \\
& PGD-AT ($\ell_2$, $\epsilon$=0.03) \citep{salman2020transfer,madry2018towards}     & 75.77                 \\
& PGD-AT ($\ell_2$, $\epsilon$=0.05) \citep{salman2020transfer,madry2018towards}     & 75.58                 \\
& PGD-AT ($\ell_2$, $\epsilon$=0.1) \citep{salman2020transfer,madry2018towards}      & 74.79                 \\
& PGD-AT ($\ell_2$, $\epsilon$=0.25) \citep{salman2020transfer,madry2018towards}     & 74.14                 \\
& PGD-AT ($\ell_2$, $\epsilon$=0.5) \citep{salman2020transfer,madry2018towards}      & 73.17                 \\
& PGD-AT ($\ell_2$, $\epsilon$=1) \citep{salman2020transfer,madry2018towards}        & 70.42                 \\
& PGD-AT ($\ell_2$, $\epsilon$=3) \citep{salman2020transfer,madry2018towards}        & 62.83                 \\
& PGD-AT ($\ell_2$, $\epsilon$=5) \citep{salman2020transfer,madry2018towards}        & 56.14                 \\
& PGD-AT ($\ell_\infty$, $\epsilon$=0.5) \citep{salman2020transfer,madry2018towards} & 73.74                 \\
& PGD-AT ($\ell_\infty$, $\epsilon$=1.0) \citep{salman2020transfer,madry2018towards} & 72.04                 \\
& PGD-AT ($\ell_\infty$, $\epsilon$=2.0) \citep{salman2020transfer,madry2018towards} & 69.09                 \\
& PGD-AT ($\ell_\infty$, $\epsilon$=4.0) \citep{salman2020transfer,madry2018towards} & 63.87                 \\
& PGD-AT ($\ell_\infty$, $\epsilon$=8.0) \citep{salman2020transfer,madry2018towards} & 54.53                 \\
\midrule
\multirow{18}{*}{\rotatebox{90}{\textbf{Augmentations}}} & AutoAugment (270Ep) \citep{Cubuk_2019_CVPRAutoAugment}                                                              & 77.50  \\
& FastAutoAugment (270Ep) \citep{Lim2019FastAutoAugment}                                                           & 77.65  \\
& RandAugment (270Ep) \citep{Cubuk2020RandAugment}                                                               & 77.64  \\
& AugMix (180Ep) \citep{hendrycks2020augmix}                                         & 77.53                 \\
& DeepAugment \citep{hendrycks2021rendition}                                         & 76.65                 \\
& DeepAugment+AugMix \citep{hendrycks2021rendition}                                  & 75.80                 \\
& Diffusion-like Noise \citep{jaini2023intriguing}                                   & 67.22                 \\
& NoisyMix \citep{erichson2022noisymix}                                              & 77.05                 \\
& OpticsAugment \citep{Muller_2023_ICCV}                                             & 74.22                 \\
& PRIME \citep{modos2022prime}                                                       & 76.91                 \\
& PixMix (180Ep) \citep{hendrycks2022pixmix}                                         & 78.09                 \\
& PixMix (90Ep) \citep{hendrycks2022pixmix}                                          & 77.36                 \\
& ShapeNet (SIN) \citep{geirhos2018imagenettrained}                                  & 60.18                 \\
& ShapeNet (SIN+IN) \citep{geirhos2018imagenettrained}                               & 74.59                 \\
& ShapeNet (SIN+IN $\rightarrow$ IN) \citep{geirhos2018imagenettrained}           & 76.72                 \\
& Texture/Shape-debiased Augmentation \citep{li2021shapetexture}                     & 76.89                 \\
& Texture/Shape-Shape Bias Augmentation \citep{li2021shapetexture}                   & 76.21                 \\
& Texture/Shape-Texture Bias Augmentation~\citep{li2021shapetexture}                 & 75.27                 \\
\midrule
\multirow{6}{*}{\rotatebox{90}{\textbf{SSL}}}                  & DINOv1 \citep{caron2021emerging}                                                  & 75.28                 \\
    & MoCo v3 (1000Ep) \citep{Chen_2021_ICCV}                                            & 74.60                 \\
    & MoCo v3 (100Ep) \citep{Chen_2021_ICCV}                                             & 68.91                 \\
    & MoCo v3 (300Ep) \citep{Chen_2021_ICCV}                                             & 72.80                 \\
    & SimCLRv2 \citep{chen2020simclrv2}                                                  & 74.90                 \\
    & SwAV \citep{caron2020swav}                                                         & 75.31                 \\
\midrule
\multirow{10}{*}{\rotatebox{90}{\textbf{Improved Training}}}     & timm A1 \citep{rw2019timm,wightman2021resnet}                                      & 80.10                 \\
     & timm A1h \citep{rw2019timm,wightman2021resnet}                                     & 80.10                 \\
 & timm A2 \citep{rw2019timm,wightman2021resnet}                                      & 79.80                 \\
 & timm A3 \citep{rw2019timm,wightman2021resnet}                                      & 77.55                 \\
 & timm B1k \citep{rw2019timm,wightman2021resnet}                                     & 79.16                 \\
 & timm B2k \citep{rw2019timm,wightman2021resnet}                                     & 79.27                 \\
 & timm C1 \citep{rw2019timm,wightman2021resnet}                                      & 79.76                 \\
 & timm C2 \citep{rw2019timm,wightman2021resnet}                                      & 79.92                 \\
 & timm D \citep{rw2019timm,wightman2021resnet}                                       & 79.89                 \\
 & TorchVision 2 \citep{BibEntry2023Nov,PyTorch}                                      & 80.34                 \\
\bottomrule
\end{tabular}
}
\end{table}

\end{document}